# Challenges in the calibration of tree-based models for imbalanced classification

Nathan Phelps[a,*], Daniel J. Lizotte[b,c,†], and Douglas G. Woolford[a,†]

[a]*Department of Statistical and Actuarial Sciences, University of Western Ontario, London, Canada*

[b]*Department of Computer Science, University of Western Ontario, London, Canada*

[c]*Department of Epidemiology and Biostatistics, University of Western Ontario, London, Canada*

**Corresponding author*

[†]*Equal contribution*

**Running title:** Calibration of tree-based models

**Email addresses:** Nathan Phelps: nphelps3@uwo.ca; Daniel J. Lizotte: dlizotte@uwo.ca; Douglas G. Woolford: dwoolfor@uwo.ca

**ORCIDs:** Nathan Phelps: 0000-0002-3173-3368; Daniel J. Lizotte: 0000-0002-9258-8619

**Acknowledgements:** We thank the Government of Alberta for providing the wildfire occurrence modelling dataset as well as B. Moore and A. Stacey for their assistance in preparing those data. We also thank the Financial Wellness Lab of the University of Western Ontario, as some of the early work for this study was done in Nathan Phelps' role as a Data Engineer with the lab.

**Funding details:** Nathan Phelps was supported by a Natural Sciences and Engineering Research Council of Canada (NSERC) Postgraduate Scholarship and an Ontario Graduate Scholarship. Daniel J. Lizotte was supported by NSERC Discovery Grant RGPIN-2024-05673. Douglas G. Woolford was supported by NSERC Discovery Grant RGPIN-2021-03920 and NSERC Strategic Network Grant NETGP 548629-19.

**Disclosure statement:** The authors report there are no competing interests to declare.

**Data availability statement:** The data used in this study was provided by Alberta Agriculture and Forestry. We are not authorized to distribute the data.

**Use of artificial intelligence statement:** ChatGPT was used to help edit the manuscript.

# Challenges in the calibration of tree-based models for imbalanced classification

When using machine learning for imbalanced binary classification problems, it is common to subsample the majority class to create a (more) balanced training dataset. This biases the model's predictions because the model learns from data that is not fully representative of the underlying population of interest. One way of accounting for this bias is analytically mapping the resulting predictions to new values based on the sampling rate for the majority class. We show that calibrating a random forest this way has negative consequences, including prevalence estimates that depend on both the number of predictors considered at each split in the random forest and the sampling rate used. We explain the former using known properties of random forests and analytical calibration and the latter by demonstrating a bias in decision trees. In contradiction with much of the existing literature, we show that decision trees can be biased towards the *minority* class. These issues indicate that tree-based models trained on undersampled data should not be calibrated analytically. Calibration approaches that can learn a miscalibration pattern in the original model (e.g., beta calibration) are more suitable.

## 1. Introduction

Severely imbalanced binary classification problems occur in many fields, with machine learning commonly used to try to solve them. If using machine learning in the presence of severe class imbalance, it is often recommended to adjust the data to generate a (more) balanced training dataset. A common way of doing so when working with large datasets is undersampling, whereby the minority class observations are retained and only a random sample of the majority class observations is kept for training. However, using this procedure biases the model's predictions because it has learned from data whose distribution differs from that of the underlying population. Let $Y$ be a binary response variable that takes a value of 1 with probability $p$ and 0 otherwise. In our study, we assume that the majority class is 0 and the minority class is 1. A perfectly calibrated model generates estimates, $\hat{p}$, such that $\mathbb{P}(Y = 1|\hat{p} = p) = p \; \forall p \in [0,1]$. A model trained on undersampled data will not be perfectly calibrated because its predicted probabilities will be upwardly biased.

Several approaches have been proposed to account for this bias, with the goal of obtaining a modelling procedure that outputs probability estimates that are well-calibrated. Multiple studies (e.g., Elkan, 2001; Saerens et al., 2002) have derived analytical solutions to account for differences between the training dataset and new data. Dal Pozzolo et al. (2015) derived a solution for the specific case where the training dataset differs from new data due to undersampling, outlining an analytical calibration method—an equation—that maps the output of a model fit to undersampled data to (hopefully) well-calibrated probabilities based on the sampling rate used for the majority class:

$$\hat{p} = \frac{\beta \hat{p}_s}{\beta \hat{p}_s - \hat{p}_s + 1} \quad \text{Eq. 1}$$

Here, $\beta$ represents the sampling rate for the majority class, $\hat{p}_s$ represents the probability estimates output by a model fit to an undersampled training dataset, and $\hat{p}$ represents the probability estimates output by the entire modelling procedure.

Dal Pozzolo et al. (2015) demonstrated the effectiveness of their method by calibrating a random forest, support vector machine, and logit boost model, and the method has since been used to calibrate a variety of machine learning models (e.g., Phelps and Woolford, 2021; Shin et al., 2023). However, we have found that this analytical calibration approach is not suitable for all machine learning models and can lead to poorly calibrated probability estimates.

In this paper, we make three key contributions. First, we demonstrate that analytical calibration is unsuitable both for calibrating the balanced random forest of Chen et al. (2004) and for calibrating a standard random forest trained on undersampled data. For the latter case, we show that calibration via Eq. 1 leads to a model whose average prediction (i.e., prevalence estimate) varies systematically with the number of predictors considered at each split in the random forest and the sampling rate used for the majority class. If the modelling procedure (particularly the calibration approach) was working as desired, neither of these hyperparameters would have an impact on prevalence estimates. Second, we explain why prevalence estimates increase with the number of predictors considered at each split, showing that this relationship is caused by properties of random forests and analytical calibration, and therefore can be expected to occur in general. Thirdly, we show that a bias in decision trees can explain why prevalence estimates change with the sampling rate. Surprisingly, we find that decision trees can be biased towards the *minority* class. While we do not find evidence that this bias occurs in all settings, this provides an important counterpoint to existing literature claiming a bias towards the majority class (e.g., Guo et al., 2008; Leevy et al., 2018; Megahed et al., 2021).

We begin by showing why analytical calibration is unsuitable for calibrating a balanced random forest. Using wildfire occurrence prediction as an example, we then demonstrate the substantial changes to prevalence estimates that can occur by changing hyperparameters when using a standard random forest. We explore these issues in more detail via a simulation study, allowing us to explain why the prevalence estimates depend on these factors. Lastly, we discuss the implications of our findings and provide conclusions and directions for future work.

## 2. Standard and balanced random forests

A random forest (Breiman, 2001) is a machine learning model composed of many (e.g., 500) decision trees. A decision tree is a low-bias model but often struggles to generalize to new data because it overfits. Random forests are designed to mitigate this issue in two ways. First, each tree is trained on a different training dataset, obtained from bootstrapping the original training dataset. Second, only a random subset of the predictors is considered at each split of each tree. The size of this random subset is a hyperparameter chosen by the modeller. When used as part of a random forest, many research articles and textbooks recommend fitting decision trees to purity (i.e., no error on their training dataset) (Zhou and Mentch, 2023).

There are different ways that undersampling can be used in conjunction with random forests. For a standard random forest implementation, undersampling is performed on the original training dataset and then bootstrapped training datasets are drawn from this new dataset. In the balanced random forest algorithm (Chen et al., 2004), the undersampling and bootstrapping processes occur simultaneously. Each training dataset is drawn from the original (unsampled) training dataset, with each dataset created from stratified bootstrapped samples of the same size from the positive and negative cases. We will demonstrate that analytical calibration via Eq. 1 is unsuitable for this algorithm.

Consider a dataset with an imbalanced binary response variable, $Y$, that takes values 0 (the majority or negative class) or 1 (the minority or positive class) and some predictor(s) denoted by $x$. Assume the dataset contains $N$ observations consisting of $N^+$ positive cases and $N^-$ negative cases. Let $S$ be a variable indicating if an observation is included in the training dataset, with $S = 1$ indicating that it is included and $S = 0$ indicating that it is excluded. Borrowing notation from Dal Pozzolo et al. (2015), we let $p = \mathbb{P}(Y = 1|x)$ and $p_s = \mathbb{P}(Y = 1|S = 1, x)$. Then from Eq. 1 in Dal Pozzolo et al. (2015), we have that $p_s = \frac{\mathbb{P}(S = 1|Y = 1)p}{\mathbb{P}(S = 1|Y = 1)p + \mathbb{P}(S = 1|Y = 0)(1-p)}$. As in Dal Pozzolo et al. (2015), consider the scenario where undersampling is done to balance a dataset without bootstrapping. In this case, $\mathbb{P}(S = 1|Y = 0) = \frac{N^+}{N^-}$. When undersampling is done with bootstrapping, we instead obtain $\mathbb{P}(S = 1|Y = 0) = 1 - \left(\frac{N^- - 1}{N^-}\right)^{N^+}$. Although these values are not equivalent, for very imbalanced datasets they are approximately equal. The main issue arises with $\mathbb{P}(S = 1|Y = 1)$. Because of the bootstrapping procedure, we no longer have $\mathbb{P}(S = 1|Y = 1) = 1$. Instead, $\mathbb{P}(S = 1|Y = 1) = 1 - \left(\frac{N^+ - 1}{N^+}\right)^{N^+} \approx 1 - e \approx 0.632$. For this reason, Eq. 1 is not appropriate for calibrating a balanced random forest.

This, however, is not a problem for calibrating a standard random forest. For a standard random forest, we still have $\mathbb{P}(S = 1|Y = 1) \approx 0.632$. However, for $\mathbb{P}(S = 1|Y = 0)$ we now have $\mathbb{P}(S = 1|Y = 0) = \frac{N^+}{N^-}\left[1 - \left(\frac{N^+-1}{N^+}\right)^{N^+}\right] \approx 0.632\,\frac{N^+}{N^-}$ because we undersample without replacement and then perform the bootstrapping procedure. Consequently, we obtain $p_s = \frac{\mathbb{P}(S = 1|Y = 1)p}{\mathbb{P}(S = 1|Y = 1)p + \mathbb{P}(S = 1|Y = 0)(1-p)} \approx \frac{0.632p}{0.632p + 0.632\frac{N^+}{N^-}(1-p)} = \frac{p}{p + \frac{N^+}{N^-}(1-p)}$. This is the same as the result from Dal Pozzolo et al. (2015).

We note that it is possible that there are issues with some observations being included twice in the training dataset of individual trees. This could influence both balanced random forests and standard random forests, but investigation of this potential issue is outside the scope of this work. Although this analysis suggests that standard random forests can be calibrated using analytical calibration, we will show in the next section that this also leads to poor probability estimates.

## 3. An illustrative example: Wildfire occurrence prediction in the Lac La Biche region of Alberta, Canada

Wildfire occurrence prediction models are important decision support tools for fire managers, as these models help them make decisions such as where to allocate limited resources (Taylor et al., 2013). On a fine spatio-temporal scale, wildfires are incredibly rare, with datasets commonly having less than 0.1% positive cases (e.g., de Haan-Ward et al., 2024). For our example, we consider a dataset from the Lac La Biche region of Alberta, Canada and focus on human-caused wildfires. This dataset was used in an earlier study that compared statistical and machine learning models for wildfire occurrence prediction (Phelps and Woolford, 2021). In the following paragraph, we provide only a cursory description of the data. For more details, see Phelps and Woolford (2021).

The dataset spans the years 1996 to 2016, with data available for the entire wildfire season (i.e., March 1 to October 31) in most years. Each voxel in the dataset (excluding those on the boundary) represents a 10km × 10km × one day region in space-time. The data for each voxel includes information on vegetation, measures of human activity such as the distance covered by roads or railways and the amount of wildland-urban interface (WUI; see Johnston and Flannigan, 2017), and weather, including both standard weather variables like temperature and relative humidity, and indices and codes from Canada's Fire Weather Index System (Van Wagner, 1987) such as the Fine Fuel Moisture Code (FFMC) and Drought Code (DC). We split the data into training and testing datasets, using the years 1996 to 2011 for training and 2012 to 2016 for testing. In the training dataset, 0.06% of the observations were wildfire occurrences.

### *3.1 Methods: Wildfire occurrence prediction*

We implemented random forests using the RandomForestClassifier function from Python's scikit-learn library (Pedregosa et al., 2011), using the default settings unless stated otherwise. Each random forest had 500 decision trees, with each of those fit to purity. To construct the training dataset for each random forest, undersampling was used. In order to demonstrate the relationship between prevalence estimates and the sampling rate, multiple sampling rates for the majority class were explored (i.e., 0.01, 0.04, 0.07, and 0.10). Models were calibrated using Eq. 1 (Dal Pozzolo et al., 2015). Each model used 15 predictors: latitude, longitude, day of year, temperature, relative humidity, FFMC, DC, Initial Spread Index (ISI), Duff Moisture Code (DMC), the distance covered by roads in the cell, the percentage of the cell covered by aspen trees, and the percentage of the cell that is water, WUI, wildland-industrial interface, and infrastructure interface. Like with the sampling rate, we needed to consider multiple values of the hyperparameter that determines the number of predictors to consider at each split in the decision trees. We considered all possible values for this hyperparameter by letting it vary from one to 15.

### *3.2 Results and discussion: Wildfire occurrence prediction*

For each modelling procedure, we computed the predicted number of wildfires in the testing dataset (i.e., from 2012 to 2016) by summing the predicted probabilities across the entire dataset. In Figure 1, we plot the predicted number of wildfires against the number of predictors considered at each split for each sampling rate. If analytical calibration was working as desired, this should result in approximately horizontal lines that overlap one another. Instead, Figure 1 shows that the range of the predicted number of wildfires is extremely large; the largest estimate is more than 40% larger than the smallest estimate. If the provincial wildfire management agency were to use one of these models to predict the number of human-caused wildfires that will occur in the Lac La Biche region, their expectations in that region would differ substantially depending on which model was used. In addition, it is clear from Figure 1 that the large range in the predicted number of wildfires is not simply due to randomness. There are clear, positive relationships between the predicted number of wildfires and both the sampling rate and the number of predictors considered at each split.

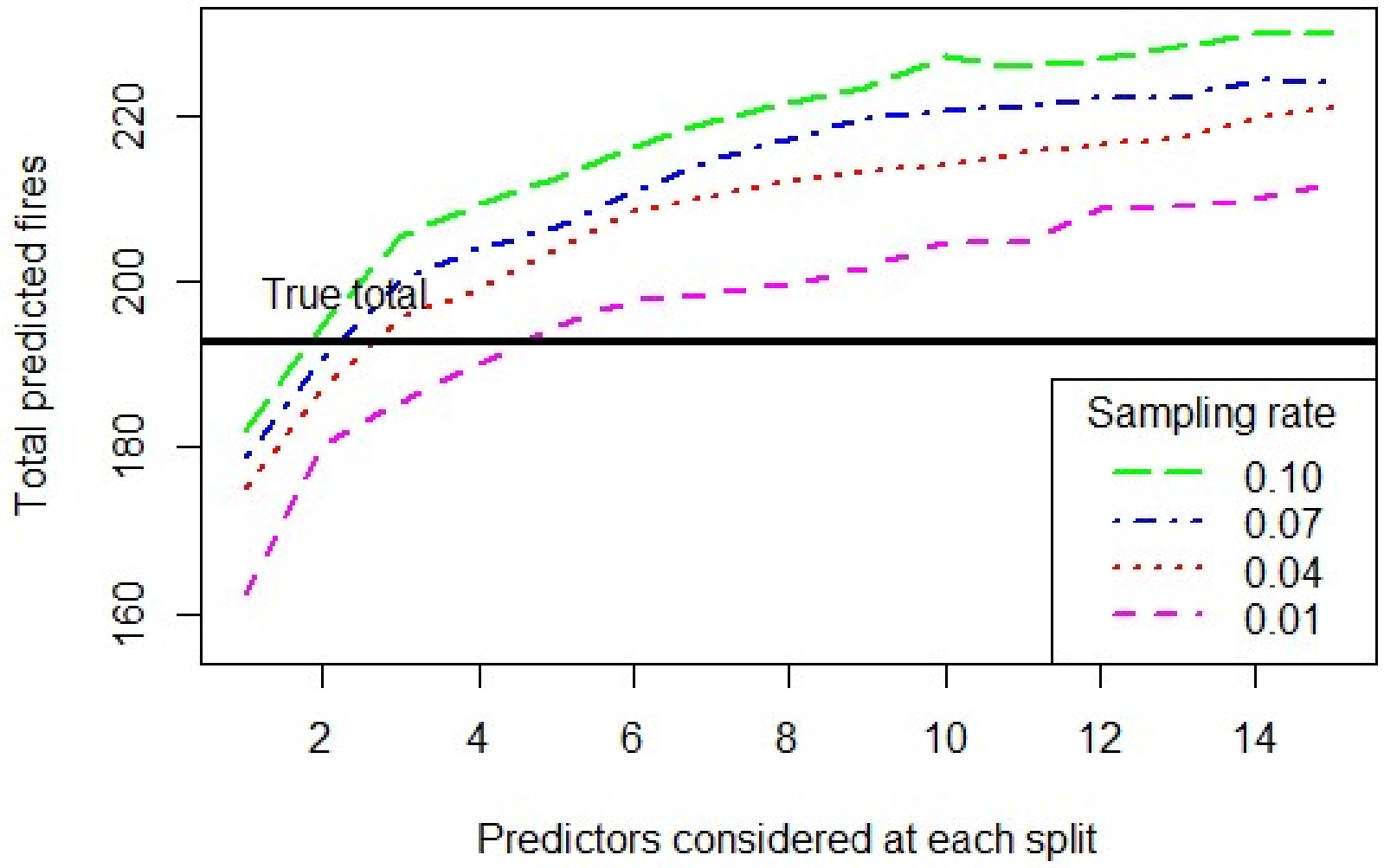

Figure 1. A plot illustrating how the predicted number of wildfires varies with the number of predictors considered at each split and the sampling rate for the majority class. The horizontal black line indicates the true number of wildfires in the testing dataset.

**4. A simulator for imbalanced data**

To investigate the relationships shown in Figure 1, we developed a simulator. This provides us with more control over the data generating process and knowledge of the true outcome probabilities. The simulator has 10 predictors, each of which follow a uniform distribution with various minimums and maximums (see Table 1).

Table 1. The minimum and maximum value for each of the 10 predictors in the simulated datasets.

| **Covariate** | **Minimum** | **Maximum** |
|---|---|---|
| 1 | -0.4 | 0.6 |
| 2 | -0.2 | 0.8 |
| 3 | -0.4 | 1 |
| 4 | -0.1 | 0.9 |
| 5 | 0 | 5 |
| 6 | 0 | 3 |
| 7 | 1 | 4 |
| 8 | 1 | 7 |
| 9 | 1 | 3 |
| 10 | 0 | 2 |

The relationship between the predictors and the outcome is determined by Eq. 2. While this relationship was created somewhat arbitrarily, it was designed to be sufficiently complicated so that it would be challenging to model using traditional statistical methods. We did this because it is in these situations that models like random forests can be especially useful.

$$\text{logit}(p) = \frac{\log(99)}{40}(x_1 + x_2 + x_3 + x_4 + x_5 + x_6 + x_7 + x_8 + x_9 + x_{10} + x_1x_3 + x_2x_5 + x_4x_9 + x_6x_7 + x_8x_{10} + x_1x_2x_3x_4 + x_1x_2x_9x_{10}) - k\log(99) \quad \text{Eq. 2}$$

In Eq. 2, $x_i$ represents the $i^{th}$ predictor and $k$ is a parameter used to adjust the level of imbalance in the data generating process. Using $k = 1.5$ (corresponding to approximately 2.08% positive cases) to generate training and testing datasets with 100 000 observations each, we verified that we obtained similar findings to those from Section 3. We used sampling rates of 0.025, 0.050, 0.075, and 0.100 and used hyperparameter values from one to 10 for the number of predictors considered at each split. We then compared the prevalence estimates obtained for the testing dataset under each modelling procedure. We repeated this process 100 times so that we could assess the consistency of our findings. The results are shown in Figure 2, which strongly resembles Figure 1.

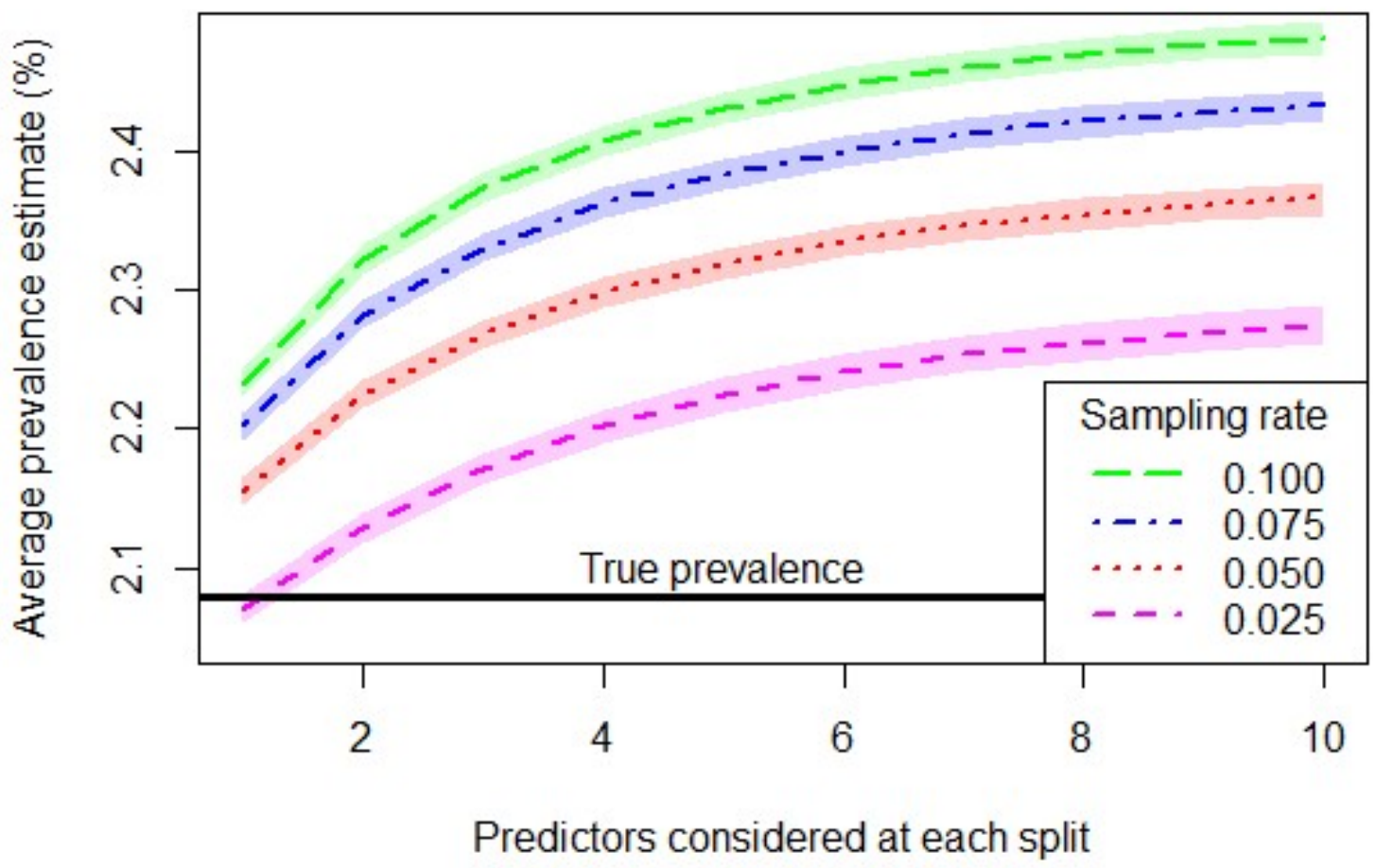

Figure 2. A plot illustrating how the prevalence estimate varies with the number of predictors considered at each split and the sampling rate. The shaded regions indicate 95% confidence intervals for the average prevalence estimate across the 100 runs. The horizontal black line represents the true prevalence from the data generating process.

As before, we find clear positive relationships between the prevalence estimates and both the sampling rate and the number of predictors considered at each split. Again, the differences in prevalence estimates can be substantial, with the maximum average prevalence estimate nearly 20% larger than the minimum. One difference between these results and those in Figure 1 is that almost all prevalence estimates now overestimate the true prevalence in our data generating process. Thus, at least in some situations, it appears that using random forests with analytical calibration can systematically lead to upwardly biased prevalence estimates.

## 5. Investigating the effect of the number of predictors considered

Sections 3 and 4 demonstrated that, with both real and simulated data, training a random forest on undersampled data and calibrating it using analytical calibration can lead to prevalence estimates that increase with the number of predictors considered at each split. When more predictors are considered, the trees in the random forest become more like one another because they are more likely to split on the same predictor. When the trees are more similar, the random forest makes more extreme predictions (i.e., closer to 0 or 1). This change in the distribution of predictions based on the number of predictors considered at each split could lead to the pattern we observed. Since Eq. 1 is non-linear, even if the prevalence estimates from random forests with a different number of predictors considered at each split were the same before calibration, the prevalence estimates might not be the same after calibration.

### *5.1 Methods: Effect of number of predictors considered*

To test our conjecture, we studied the distribution of the random forest's predictions when two and 10 predictors were considered at each split. The distributions of the predictions before and after calibration were compared using quantile-quantile plots. We also paid special attention to the minimum and maximum predictions from each model before and after calibration. These random forests were both fit to data simulated from the data generating process described in Section 4, with a sampling rate of 0.03. Because this process does not involve running the simulation several times, we used training and testing datasets with one million observations.

### *5.2 Results and discussion: Effect of number of predictors considered*

Prior to using Eq. 1 to calibrate the predictions of the two models, the prevalence estimates (i.e., average predictions) from the two models were virtually identical. This is unsurprising considering the very similar distributions shown in the left plot in Figure 3. As expected, the model considering 10 predictors at each split made more extreme predictions, but the difference in predictions was very small. After calibration, however, the distributions of the predictions differed substantially (see right plot in Figure 3). Although the distributions of the predictions still seem similar for small probability estimates, there is a large deviation from the 45° line for larger probability estimates. The prevalence estimates from the two models also differed after calibration; the average prediction from the model considering 10 predictors at each split was 4.9% larger than the average prediction from the model considering only two predictors.

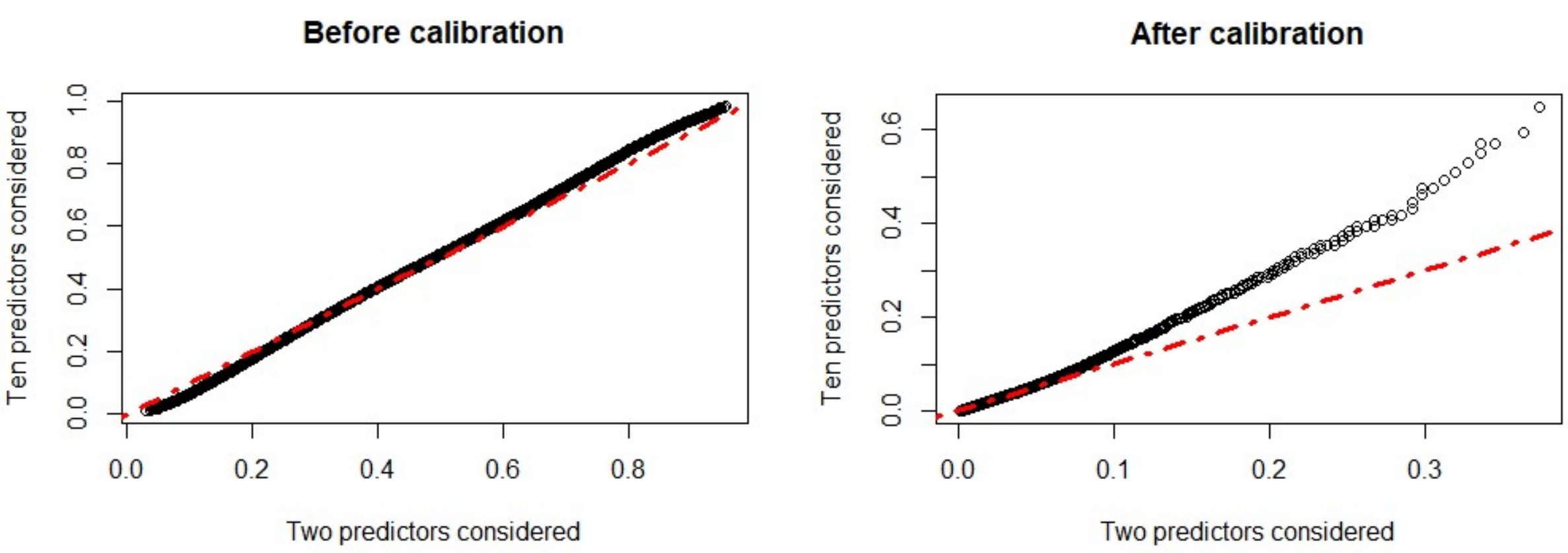

Figure 3. Quantile-quantile plots comparing the distributions of predictions from random forest models that considered two predictors at each split and 10 predictors at each split, before (left) and after (right) calibration to account for undersampling. The dashed line is the 45° line.

Comparing the maximum and minimum predictions from each random forest before and after calibration illustrates why the prevalence estimates differ after calibration. For the random forest with all 10 predictors considered at each split, the maximum prediction was 0.984 before calibration and 0.649 after calibration. For the random forest considering two predictors, the maximum prediction was 0.952 before calibration and 0.373 after calibration. This means the ratio between maximum predictions changed from 1.034 to 1.738. The minimum predictions for the model considering 10 predictors at each split were 0.012 and $3.64\times10^{-4}$ before and after calibration, respectively. The corresponding minimum predictions for the model considering two predictors were 0.030 and $9.27\times10^{-4}$, meaning the ratio between the minimum predictions remained virtually unchanged, going from 0.400 before calibration to 0.393 after calibration.

This phenomenon can be explained by reframing how we think of analytical calibration. For a given sampling rate, we can compute a multiplicative adjustment factor to apply to each original prediction that will map them to the same values as Eq. 1. For the sampling rate of 0.03 used here, we show the multiplicative adjustment factor for each original prediction in Figure 4. For very small original predicted probabilities, the adjustment factor is virtually constant, explaining why the ratio between the minimum predictions is nearly identical before and after calibration. However, for larger predicted probabilities, the adjustment factor differs dramatically, altering the ratio for the maximum predictions. Combined with the slightly more extreme predictions from models considering more predictors at each split, it is this property of analytical calibration that causes prevalence estimates to increase as the number of predictors considered at each split increases. Given that this phenomenon relies only on a well-known property of random forests and the non-linear nature of analytical calibration, we have no reason to believe that it would not occur in general (i.e., when modelling other datasets) if analytical calibration is used to calibrate a random forest trained on undersampled data.

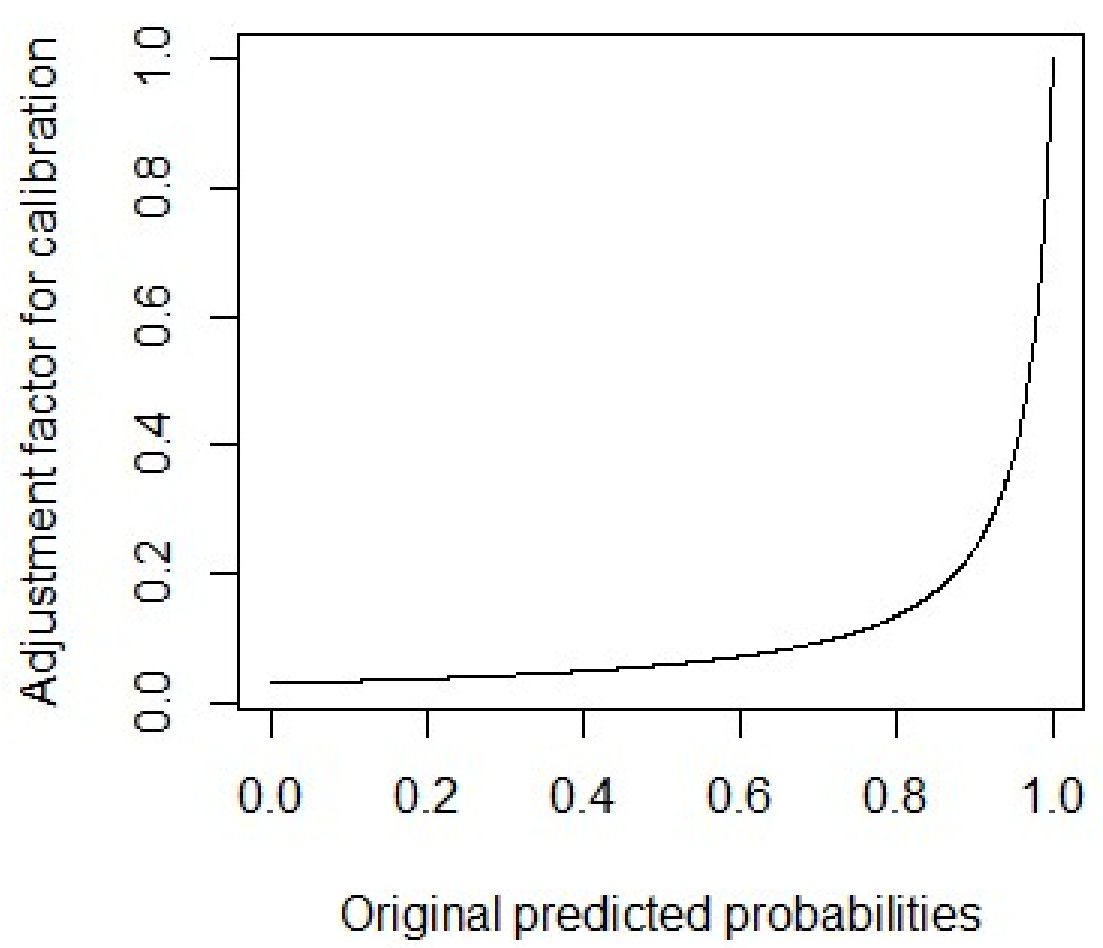

Figure 4. A plot showing a multiplicative adjustment factor that can be applied to original predicted probabilities to obtain the same post-calibration predicted probabilities as obtained from Eq. 1 when the sampling rate is 0.03.

## 6. Investigating the effect of the sampling rate

Our results in Sections 3 and 4 showed prevalence estimates that increased with the sampling rate for the majority class. Since increasing the sampling rate increases the level of class imbalance in the training dataset, a possible explanation for this relationship is that decision trees are biased towards the minority class when learning from imbalanced data. In this case, the bias we are referring to means that the random forest inflates probability estimates. Thus, this conjecture is also consistent with the systematic overprediction observed in Figure 2. If the decision trees composing the random forests are overpredicting, the random forests will overpredict too; analytical calibration will not account for this.

### *6.1 Methods: Effect of sampling rate*

To test if decision trees are biased towards the minority class or not, we used our data generating process from Section 4. We generated datasets with varying levels of class imbalance, ranging from nearly balanced (49.8% positive cases) to extremely imbalanced (0.2% positive cases). Both training and testing datasets had one million observations. We fit individual decision trees to these datasets, with each tree fit to purity because we were interested in their performance in the context of random forests. It is worth noting that these decision trees were still implemented using the RandomForestClassifier function, just without bootstrapping, with only a single tree, and with all predictors considered at each split. We did this because we found that we sometimes obtained slightly different results when using the DecisionTreeClassifier function. For each level of class imbalance, we computed the ratio of the mean prediction on a corresponding testing dataset to the mean true probability in that dataset. We ran this process 100 times and computed the mean and standard deviation of the ratio for each level of class imbalance.

### *6.2 Results and discussion: Effect of sampling rate*

To explain the increasing prevalence estimates as the sampling rate increases, we investigated potential overprediction by decision trees in imbalanced data. The results of this investigation are shown in Table 2. Clearly, our results show that the decision tree tended to overpredict when learning from imbalanced data. Except for when the dataset was nearly balanced (49.8% positive cases), we have overwhelming evidence suggesting that the ratio between the decision tree's predictions and true outcome probabilities is larger than one. In some cases, the magnitude of the overprediction is quite large, reaching as high as 23.4%. The magnitude of the overprediction generally increased as the level of imbalance increased, but this trend did not continue for the most extreme level of class imbalance. However, the overprediction for this dataset was still substantial and the standard deviation for this ratio is large, so it is plausible the reduction in overprediction is due to randomness. As an additional robustness check to ensure our results were not dependent on predictors following a uniform distribution, we implemented a similar experiment with normally distributed and lognormally distributed predictors. Details are provided in Appendix 1. The results confirmed that the overprediction problem is not specific to uniform predictors, but the magnitude of the overprediction did change.

Table 2. The average and standard deviation (SD) of the ratio of the mean predicted probability on the testing dataset to mean true probability in the testing dataset for a decision tree fit to datasets with varying levels of class imbalance based on Eq. 2.

| **Average prevalence** | 0.498 | 0.397 | 0.305 | 0.225 | 0.160 | 0.061 | 0.021 | 0.002 |
|---|---|---|---|---|---|---|---|---|
| **Ratio of predictions to success probabilities in the testing dataset** | 1.001 | 1.019 | 1.044 | 1.073 | 1.106 | 1.188 | 1.233 | 1.157 |
| **SD of ratio of predictions to success probabilities in the testing dataset ($\times 10^{-3}$)** | 1.458 | 1.705 | 2.312 | 3.039 | 3.948 | 6.905 | 11.166 | 43.866 |

Our finding that decision trees can be biased towards the minority class is inconsistent with the widespread belief that machine learning models, including decision trees, "underestimate" (Megahed et al., 2021), "ignore" (Guo et al., 2008), or "neglect" (Japkowicz and Stephen, 2002) the minority class. There is certainly more work to be done to understand and explain these differing results. With that said, we believe that we have provided very compelling evidence of a bias towards the minority class in the setting we have considered. Thus, although it may not always be true that decision trees are biased towards the minority class, we have clearly shown that this can be the case. This alone is a very surprising finding.

## 7. Discussion

In this study, we have illustrated problems with using analytical calibration (Dal Pozzolo et al., 2015) to calibrate random forests trained on undersampled datasets. For balanced random forests (Chen et al., 2004), we showed that analytical calibration via Eq. 1 is unsuitable because the probabilities used to derive that equation do not apply. For standard random forests, we illustrated that, when modelling wildfire occurrences, the expected number of wildfires changed considerably when different seemingly reasonable values were chosen for the number of predictors considered at each split and the sampling rate. In the simulation study, we obtained estimates that also appear systematically upwardly biased. Wildfire management agencies allocate their resources based in part on the expected number of fires in various regions, so systematically erroneous estimates would hinder their ability to use their resources effectively. It is clear from these results that analytical calibration of random forests has failed to provide well-calibrated probability estimates; good prevalence estimates are a necessary (but insufficient) condition for good calibration. Thus, if using a random forest to learn from undersampled data, an alternative calibration approach should be used. We expect calibration approaches that can learn to adjust for the miscalibration of random forests will be more effective, such as beta calibration (Kull et al., 2017) or modified versions of Platt's scaling (Phelps et al., 2025). It may be worthwhile for future studies to evaluate the effectiveness of these approaches for calibrating random forests trained on undersampled data.

In addition to identifying problems with analytically calibrating a random forest after undersampling, we have explained why these problems exist. First, we explained the relationship between the number of predictors considered at each split and prevalence estimates using known properties of random forests and analytical calibration. Random forests make more extreme predictions when more predictors are considered at each split. Typically, this property of random forests makes very little difference in the predictions (e.g., see the left side of Figure 3). However, in conjunction with the non-linear nature of Eq. 1, this results in materially different predictions and, ultimately, prevalence estimates. Second, we explained the positive relationship between sampling rates and prevalence estimates with the surprising finding that decision trees can be biased towards the minority class when modelling imbalanced data.

When modelling wildfire occurrences, we considered other configurations in addition to those presented in Section 3. We also considered an R implementation of random forests using the randomForest package (Liaw and Wiener, 2002), a sampling rate of 0.001, and versions of the models with only 11 predictors. These additional results are shown in Appendix 2, but we summarize the key findings here. In all configurations, we found the positive relationship between prevalence estimates and the number of predictors considered at each split. This aligns with our expectation that this phenomenon will occur in general. However, the remaining results were far less consistent with the results shown in Figure 1. First, the R implementation provided meaningfully different estimates from the Python implementation. In addition, the relationship between prevalence estimates and the sampling rate was dependent on the number of predictors considered at each split. We found that the R implementation did not behave as expected, with trees not quite fit to purity. However, it is not clear if this is the cause of the different outcomes relative to Python. Second, using a sampling rate of 0.001 led to some very extreme predictions, with estimates sometimes exceeding two times the observed number of fires. It is also not clear at this time why this occurred, but this should encourage caution when sampling only an extremely small proportion of the majority class (even if the resulting dataset still contains more negative cases than positive cases). Thirdly, the relationship between prevalence estimates and the sampling rate changed for the Python implementation when only 11 predictors were used in the model. Again, it is not clear why such a change would result in changes to this relationship.

Although we did not expect a priori that there is a relationship between prevalence estimates and the number of predictors considered at each split, in hindsight it is unsurprising that we consistently found this relationship. Now knowing that this relationship is caused by a well-established property of random forests (i.e., they make more extreme predictions when more predictors are considered at each split) and the non-linearity of analytical calibration, consistently finding this relationship is the expected outcome.

On the contrary, the finding that decision trees can be biased towards the minority class contradicts a large body of literature. Broadly speaking, machine learning models are thought to struggle to learn from imbalanced data because they are biased towards the majority class (e.g., Guo et al., 2008; Leevy et al., 2018; Megahed et al., 2021). Decision trees are no exception, as Japkowicz and Stephen (2002) found that decision trees "neglect the minority class", while several other studies have focused on addressing issues with class imbalance specifically for decision trees (e.g., Boonchuay et al., 2017; Cieslak and Chawla, 2008; Liu et al., 2010; Prati et al., 2008). However, Plante and Radatz's (2024) simulation study found biases in decision trees and random forests in both directions on different datasets (although some biases were very small and therefore possibly due to chance). We believe we have shown compelling evidence of a bias towards the minority class in the simulation settings we considered. Thus, a bias towards the minority class is our explanation for the positive relationship between prevalence estimates and the sampling rate. However, this relationship was not observed consistently outside the simulation settings. Thus, our own results—through the analysis of both real and simulated data—indicate this bias towards the minority class does not always occur and that the bias can go in the other direction. However, considering all the existing literature on biases in decision trees, it is an important finding that this bias towards the minority class can occur in realistic, practical settings. When faced with imbalanced data, one should not automatically assume that decision trees will be biased towards the majority class.

Herein, we have exclusively focused on trees that are fit to purity. While this is commonplace in random forests, recent work has advocated for tuning tree depth rather than fitting to purity (Zhou and Mentch, 2023). Future work could consider studying the use of analytical calibration with random forests whose decision trees are not fit to purity. Other valuable future work could include developing explanations for the unexplained behavior in Appendix 2, such as the differences between Python and R implementations and the inconsistent relationship between prevalence estimates and sampling rates. It will also be important to develop a better understanding of the biases in decision trees. Overall, there is clearly still substantial work to be done in order to fully understand the phenomena we have observed.

## 8. Conclusion

We have demonstrated that using the equation of Dal Pozzolo et al. (2015) to calibrate random forests trained on undersampled datasets generates prevalence estimates that are dependent on both the number of predictors considered at each split in the random forest and the sampling rate used. It is important to make this issue known, as random forests were used to demonstrate the efficacy of analytical calibration (Dal Pozzolo et al., 2015), the procedure has been implemented by others (Shin et al., 2023), and the impact on end users can be substantial. Rather than using analytical calibration, modellers should use calibration methods that can learn to account for the miscalibration of random forests in addition to the miscalibration caused by undersampling.

We have also explained the causes behind the relationships we observed with the prevalence estimates. The relationship with the number of predictors considered at each split occurs because analytical calibration non-linearly adjusts the original predictions, which are more extreme when random forests consider more predictors at each split. The relationship with the sampling rate is driven by a bias in decision trees. Importantly, we found that decision trees can be biased towards the minority class, a very surprising finding considering the large body of literature that claims decision trees are biased in the opposite direction. Given that this finding has implications for popular models like random forests, as well as other tree-based models, it will be important for future work to try to address this discrepancy.

## Appendix 1

To ensure that the bias towards the minority class that we observed in Table 2 was not because the predictors were uniformly distributed, we repeated the simulation study described in Section 6 with normal and lognormal predictors. Parameters for each covariate are provided in Table A1.

Table A1. The parameters for each predictor, $X_i$, used to create the simulated datasets.

| **Covariate** | **Normal predictors** | | **Lognormal predictors** | |
|---|---|---|---|---|
| | Mean of $X_i$ | Standard deviation of $X_i$ | Mean of log of $X_i$ | Standard deviation of log of $X_i$ |
| 1 | 0.5 | 0.5 | 0.05 | 0.05 |
| 2 | 0.5 | 0.8 | 0.05 | 0.08 |
| 3 | -0.2 | 1 | -0.02 | 0.1 |
| 4 | -0.1 | 0.9 | -0.01 | 0.09 |
| 5 | 0 | 5 | 0.2 | 0.5 |
| 6 | 0 | 3 | 0 | 0.3 |
| 7 | 2 | 4 | 0.2 | 0.4 |
| 8 | 3 | 7 | 0.3 | 0.7 |
| 9 | 1.5 | 3 | 0.15 | 0.3 |
| 10 | 0 | 2 | 0 | 0.2 |

As shown in Table A2, we considered five different prevalences for each distribution of the predictors. The results show that the overprediction issue persisted when predictors did not follow a uniform distribution, although the magnitude of the overprediction was substantially reduced with the normally distributed predictors.

Table A2. The average and standard deviation (in parentheses) of the ratio of the mean predicted probability on the testing dataset to mean true probability in the testing dataset for a decision tree fit to datasets with varying levels of class imbalance based on Eq. 2.

| **Normal predictors** | | **Lognormal predictors** | |
|---|---|---|---|
| Prevalence | Ratio of predictions to success probabilities in the testing dataset | Prevalence | Ratio of predictions to success probabilities in the testing dataset |
| 0.465 | 1.004 (0.001) | 0.419 | 1.016 (0.002) |
| 0.257 | 1.022 (0.002) | 0.230 | 1.079 (0.003) |
| 0.126 | 1.034 (0.003) | 0.110 | 1.157 (0.005) |
| 0.016 | 1.037 (0.010) | 0.020 | 1.231 (0.013) |
| 0.007 | 1.036 (0.017) | 0.001 | 1.091 (0.051) |

**Appendix 2**

Here, we show additional results from the other configurations used for wildfire occurrence prediction. In Figure A1, we provide a comparison of the Python and R implementations of the model described in Section 3 (i.e., with all 15 predictors) with sampling rates of 0.01, 0.04, 0.07, and 0.10. For both implementations, there is a clear positive relationship between the predicted number of fires and the number of predictors considered at each split. The main differences between the two implementations arise when the number of predictors considered at each split is small. In these cases, the R implementation predicts far fewer fires relative to both the Python implementation and the actual number of fires. In addition, for the R implementation, when only a small number of predictors is considered at each split, it seems that prevalence estimates generally decrease as the sampling rate increases. However, when more predictors are considered at each split, this relationship reverses and resembles the relationship seen in the Python implementation.

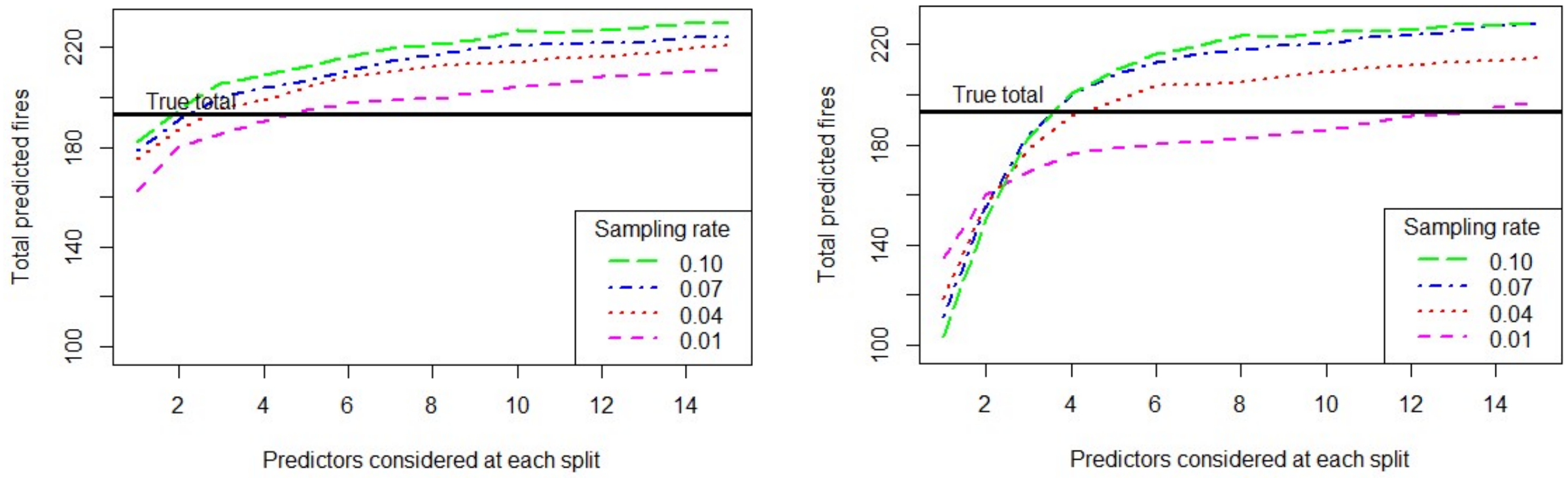

Figure A1. Plots illustrating how the predicted number of wildfires varies with the number of predictors considered at each split and the sampling rate. The left and right plots are for an implementation in Python and R, respectively. The horizontal black line indicates the true number of wildfires in the testing dataset.

The model with 11 predictors uses a subset of the 15 predictors described in Section 3. It uses latitude, longitude, day of year, Fine Fuel Moisture Code (FFMC), Duff Moisture Code (DMC), the distance covered by roads in the cell, the percentage of the cell covered by aspen trees, and the percentage of the cell that is water, wildland-urban interface (WUI), wildland-industrial interface, and infrastructure interface. The results from using this model, implemented in either Python or R, are shown in Figure A2. Like in Figure A1, the predictions from the two implementations differ dramatically when smaller subsets of the predictors are considered at each split. For the R implementation, the patterns look very similar to those observed in Figure 1. For the Python implementation, however, the relationship between the predicted number of fires and the sampling rate seems to be in the opposite direction.

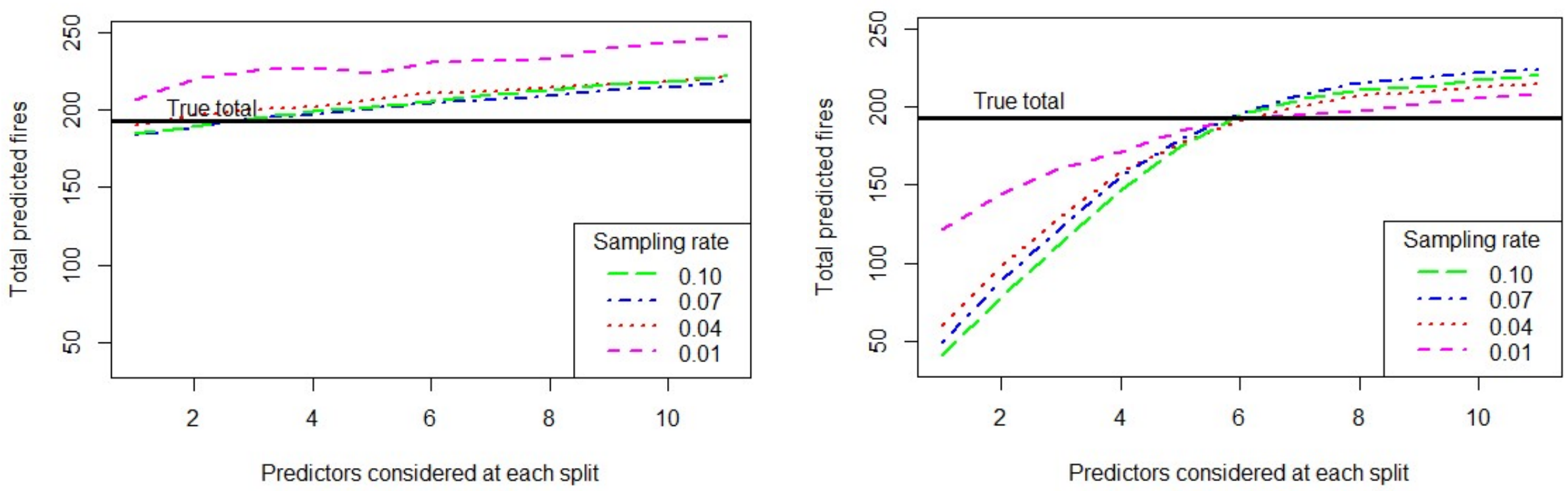

Figure A2. Plots illustrating how the predicted number of wildfires varies with the number of predictors considered at each split and the sampling rate. The left and right plots are for an implementation in Python and R, respectively. The horizontal black line indicates the true number of wildfires in the testing dataset.

Finally, Figure A3 shows the results for all four models when using a sampling rate of 0.001. Although this is an extremely small sampling rate, this still generates datasets with more negative cases than positive cases due to the level of class imbalance in the dataset. These results were separated from the others because this sampling rate led to some very extreme predictions, including at times predictions of more than twice as many fires as observed in the testing dataset.

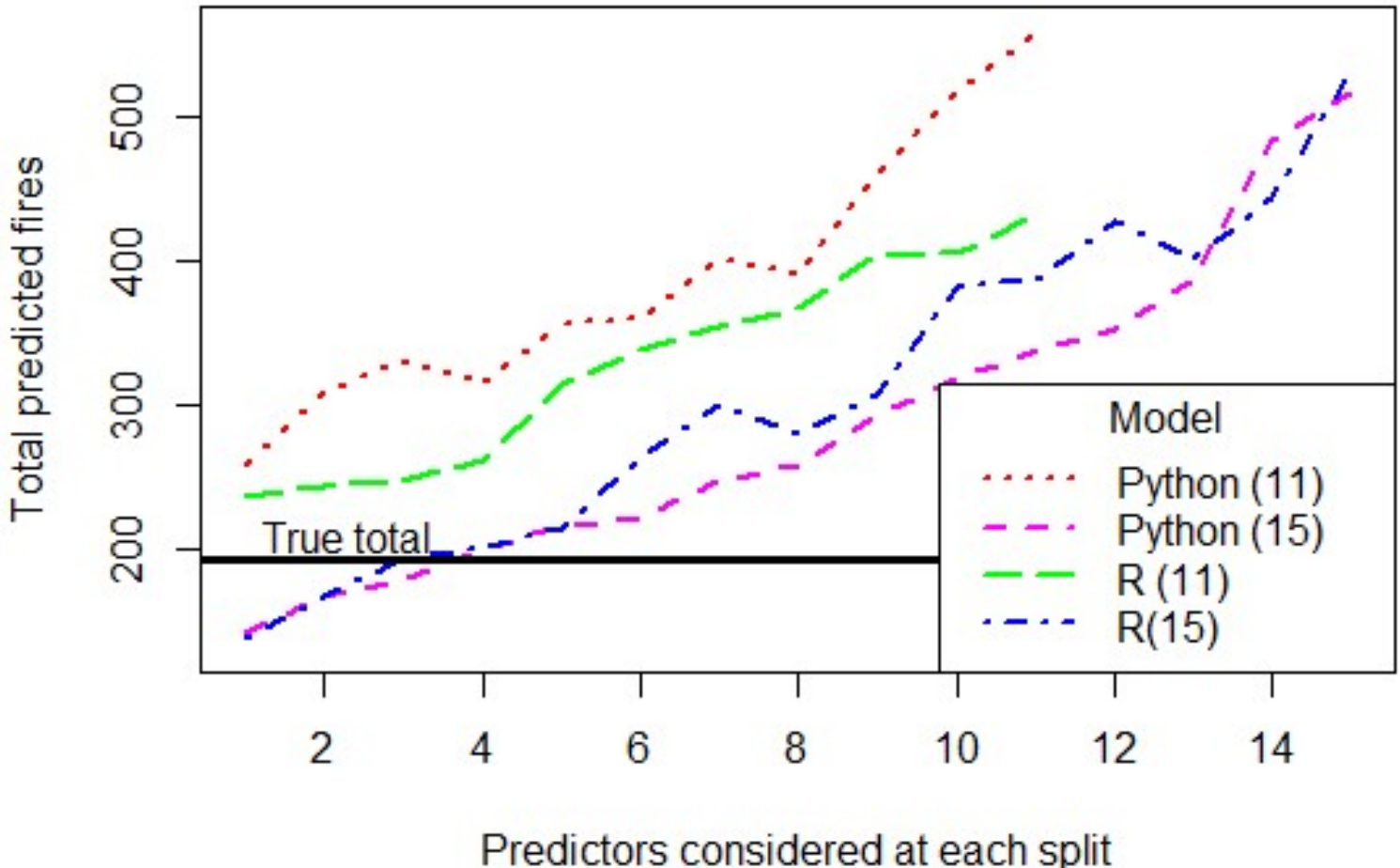

Figure A3. A plot illustrating how the predicted number of wildfires varies with the number of predictors considered at each split for different random forest models, all trained on an undersampled dataset created using a sampling rate of 0.001 for the majority class. The horizontal black line indicates the true number of wildfires in the testing dataset and the numbers in parentheses in the legend indicate the number of predictors used in the model.